# A Freely Available Morphological Analyzer, Disambiguator and Context Sensitive Lemmatizer for German[1]

Wolfgang Lezius
University of Paderborn
Cognitive Psychology
D-33098 Paderborn
lezius@psycho.uni-paderborn.de

Reinhard Rapp
University of Mainz
Faculty of Applied Linguistics
D-76711 Germersheim
rapp@usun1.fask.uni-mainz.de

Manfred Wettler
University of Paderborn
Cognitive Psychology
D-33098 Paderborn
wettler@psycho.uni-paderborn.de

**Abstract**

In this paper we present Morphy, an integrated tool for German morphology, part-of-speech tagging and context-sensitive lemmatization. Its large lexicon of more than 320,000 word forms plus its ability to process German compound nouns guarantee a wide morphological coverage. Syntactic ambiguities can be resolved with a standard statistical part-of-speech tagger. By using the output of the tagger, the lemmatizer can determine the correct root even for ambiguous word forms. The complete package is freely available and can be downloaded from the World Wide Web.

## Introduction

Morphological analysis is the basis for many NLP applications, including syntax parsing, machine translation and automatic indexing. However, most morphology systems are components of commercial products. Often, as for example in machine translation, these systems are presented as black boxes, with the morphological analysis only used internally. This makes them unsuitable for research purposes. To our knowledge, the only wide coverage morphological lexicon readily available is for the English language (Karp, Schabes, et al., 1992). There have been attempts to provide free morphological analyzers to the research community for other languages, for example in the MULTEXT project (Armstrong, Russell, et al., 1995), which developed linguistic tools for six European languages. However, the lexicons provided are rather small for most languages. In the case of German, we hope to significantly improve this situation with the development of a new version of our morphological analyzer Morphy.

In addition to the morphological analyzer, Morphy includes a statistical part-of-speech tagger and a context-sensitive lemmatizer. It can be downloaded from our web site as a complete package including documentation and lexicon (http://www-psycho.uni-paderborn.de/lezius/).
The lexicon comprises 324,000 word forms based on 50,500 stems. Its completeness has been checked using *Wahrig Deutsches Wörterbuch*, a standard dictionary of German (Wahrig, 1997). Since Morphy is intended not only for linguists, but also for second language learners of German, the current version has been implemented with Delphi for a standard Windows 95 or Windows NT platform and great effort has been put in making it as user friendly as possible. For UNIX users, an export facility is provided which allows generating a lexicon of full forms together with their morphological descriptions in text format.

## 1 The Morphology System

Since German is a highly inflectional language, the morphological algorithms used in Morphy are rather complex and can not be described here in detail (see Lezius, 1996). In essence, Morphy is a computer implementation of the morphological system described in the Duden grammar (Drosdowsky, 1984).

An overview on other German morphology systems, namely *GERTWOL, LA-Morph, Morph, Morphix, Morphy, MPRO, PC-Kimmo*

---

and *Plain,* is given in the documentation for the Morpholympics (Hausser, 1996). The Morpholympics were an attempt to compare and evaluate morphology systems in a standardized competition. Since then, many of the systems have been further developed. The version of Morphy as described here is a new release. Improvements over the old version include an integrated part-of-speech tagger, a context-sensitive lemmatizer, a 2.5 times larger lexicon and more user-friendliness through an interactive Windows-environment.

The following subsections describe the three submodules of the morphological analyzer. These are the lexical system, the generation module and the analysis module.

## 1.1 Lexical System

The lexicon of Morphy is very compact as it only stores the base form for each word together with its inflection class. Therefore, the complete morphological information for 324,000 word forms takes less than 2 Megabytes of disk space. In comparison, the text representation of the same lexicon, which can be generated via Morphy´s export facility, requires 125 MB when full morphological descriptions are given.

Since the lexical system has been specifically designed to allow a user-friendly extension of the lexicon, new words can be added easily. To our knowledge, Morphy is the only morphology system for German whose lexicon can be expanded by users who have no specialist knowledge. When entering a new word, the user is asked the minimal number of questions necessary to infer the grammatical features of the new word and which any native speaker of German should be able to answer.

## 1.2 Generation

Starting from the root form of a word and its inflection type as stored in the lexicon, the generation system produces all inflected forms. Morphy's generation algorithms were designed with the aim of producing 100% correct output. Among other morphological characteristics, the algorithms consider vowel mutation (Haus - Häuser), shift between ß and ss (Faß - Fässer), e-omission (segeln - segle), infixation of infinitive markers (weggehen - weg<u>zu</u>gehen), as well as pre- and infixation of markers of participles (gehen - <u>ge</u>gangen; weggehen - weg<u>ge</u>gangen).

## 1.3 Analysis

For each word form of a text, the analysis system determines its root, part of speech, and - if appropriate - its gender, case, number, person, tense, and comparative degree. It also segments compound nouns using a longest-matching rule which works from right to left and takes linking letters into account. To compound German nouns is not trivial: it can involve base forms and/or inflected forms (e.g. *Haus-meister* but *Häuser-meer*); in some cases the compounding is morphologically ambiguous (e.g. *Stau-becken* means *water reservoir*, but *Staub-ecken* means *dust corners*); and the linking letters *e* and *s* are not always determined phonologically, but in some cases simply occur by convention (e.g. *Schwein-e-bauch* but *Schwein-s-blase* and *Schwein-kram*).

Since the analysis system treats each word separately, ambiguities can not be resolved at this stage. For ambiguous word forms, all possible lemmata and their morphological descriptions are given (see Table 1 for the example *Winde*). If a word form can not be recognized, its part of speech is predicted by a guesser which makes use of statistical data derived from German suffix frequencies (Rapp, 1996).

| *morphological description* | | | | *lemma* |
|---|---|---|---|---|
| SUB | NOM | SIN | FEM | Winde |
| SUB | GEN | SIN | FEM | Winde |
| SUB | DAT | SIN | FEM | Winde |
| SUB | AKK | SIN | FEM | Winde |
| SUB | DAT | SIN | MAS | Wind |
| SUB | NOM | PLU | MAS | Wind |
| SUB | GEN | PLU | MAS | Wind |
| SUB | AKK | PLU | MAS | Wind |
| VER | SIN | 1PE | PRÄ | winden |
| VER | SIN | 1PE | KJ1 | winden |
| VER | SIN | 3PE | KJ1 | winden |
| VER | SIN | IMP | | winden |

Table 1: Morphological analysis for *Winde*.

Morphy's algorithm for analysis is motivated by linguistic considerations. When analyzing a word form, Morphy first builds up a list of possible roots by cutting off all possible prefixes and suffixes and reverses the process of vowel mutation if umlauts are found (shifts between ß and ss are treated analogously). Each root is looked up in the lexicon, and - if found - all possible inflected forms are generated. Only those roots which lead to an inflected form identical to the original word form are selected (Lezius, 1996).

Naturally, this procedure is much slower than a simple algorithm for the lookup of word forms in a full form lexicon. It results in an analysis speed of about 300 word forms per second on a fast PC, compared to many thousands using a full form lexicon. However, there are also advantages: First, as mentioned above, the lexicon can be kept very small, which is an important consideration for a PC-based system intended for Internet-distribution. More importantly, the processing of German compound nouns and the implementation of derivation rules - although only partially completed at this stage - fits better into this concept. For the processing of very large corpora under UNIX, we have implemented a lookup algorithm which operates on the Morphy-generated full form lexicon.

The coverage of the current version of Morphy was evaluated with the same test corpus that had been used at the Morpholympics. This corpus comprises about 7.800 word forms in total and consists of two political speeches, a fragment of the LIMAS-corpus, and a list of special word forms. The present version of Morphy recognized 94.3%, 98.4%, 96.2%, and 88.9% of the word forms respectively. The corresponding values for the old version of Morphy, with a 2.5 times smaller lexicon, had been 89.2%, 95.9%, 86.9%, and 75.8%.

## 2 The Disambiguator

Since the morphology system only looks at isolated word forms, words with more than one reading can not be disambiguated. This is done by the disambiguator or tagger, which takes context into account by considering the conditional probabilities of tag sequences. For example, in the sentence *"he opens the can"* the verb-reading of *can* may be ruled out because a verb can not follow an article.

After the success of statistical part-of-speech taggers for English, there have been quite a few attempts to apply the same methods to German. Lezius, Rapp & Wettler (1996) give an overview on some German tagging projects. Although we considered a number of algorithms, we decided to use the trigram algorithm described by Church (1988) for tagging. It is simple, fast, robust, and - among the statistical taggers - still more or less unsurpassed in terms of accuracy.

Conceptually, the Church-algorithm works as follows: For each sentence of a text, it generates all possible assignments of part-of-speech tags to words. It then selects that assignment which optimizes the product of the lexical and contextual probabilities. The lexical probability for word N is the probability of observing part of speech X given the (possibly ambiguous) word N. The contextual probability for tag Z is the probability of observing part of speech Z given the preceding two parts of speech X and Y. It is estimated by dividing the trigram frequency XYZ by the bigram frequency XY. In practice, computational limitations do not allow the enumeration of all possible assignments for long sentences, and smoothing is required for infrequent events. This is described in more detail in the original publication (Church, 1988).

Although more sophisticated algorithms for unsupervised learning - which can be trained on plain text instead on manually tagged corpora - are well established (see e.g. Merialdo, 1994), we decided not to use them. The main reason is that with large tag sets, the sparse-data-problem can become so severe that unsupervised training easily ends up in local minima, which can lead to poor results without any indication to the user. More recently, in contrast to the statistical taggers, rule-based tagging algorithms have been suggested which were shown to reduce the error rate significantly (Samuelsson & Voutilainen, 1997). We consider this a promising approach and have started to develop such a system for German with the intention of later inclusion into Morphy.

The tag set of Morphy's tagger is based on the feature system of the morphological analyzer. However, some features were discarded for tagging. For example, the tense of verbs is not considered. This results in a set of about 1000 different tags. A fragment of 20,000 words from the *Frankfurter Rundschau Corpus*, which we have been collecting since 1992, was tagged with this tag set by manually selecting the correct choice from the set of possibilities gener-ated by the morphological analyzer. In the following we refer to this corpus as the training corpus. Of all possible tags, only 456 actually occurred in the training corpus. The average ambiguity rate was 5.4 tags per word form.

The performance of our tagger was evaluated by running it on a 5000-word test sample of the Frankfurter Rundschau-Corpus which was distinct from the training text. We also tagged the test sample manually and compared the results. 84.7% of the tags were correctly tagged. Although this result may seem poor at first glance, it should be noted that the large tag sets have many fine distinctions which lead to a high error rate. If a tag set does not have these distinctions, the accuracy improves significantly. In order to show this, in another experiment we mapped our large tag set to a smaller set of 51 tags, which is comparable to the tag set used in the Brown Corpus (Greene & Rubin, 1971). As a result, the average ambiguity rate per word decreased from 5.4 to 1.6, and the accuracy improved to 95.9%, which is similar to the accuracy rates reported for statistical taggers with small tag sets in various other languages. Table 2 shows a tagging example for the large and the small tag set.

| *Word* | *large tag set* | *small tag set* |
| Ich | PRO PER NOM SIN 1PE | PRO PER |
| meine | VER 1PE SIN | VER |
| meine | POS AKK SIN FEM ATT | POS ATT |
| Frau | SUB AKK FEM SIN | SUB |
| . | SZE | SZE |

Table 2: Tagging example for both tag sets.

## 3 The Lemmatizer

For lemmatization (the reduction to base form), the integrated design of Morphy turned out to be advantageous. In the first step, the morphology-module delivers all possible lemmata for each word form. Secondly, the tagger determines the grammatical categories of the word forms. If, for any of the lemmata, the inflected form corresponding to the word form in the text does not agree with this grammatical category, the respective lemma is discarded. For example, in the sentence *"ich meine meine Frau"* ("I mean my wife"), the assignment of the two middle words to the verb *meinen* and the possessive pronoun *mein* is not clear to the morphology system.

However, since the tagger assigns the tag sequence *"pronoun verb pronoun noun"* to this sentence, it can be concluded that the first occurrence of *meine* must refer to the verb *meinen* and the second to the pronoun *mein*.

Unfortunately, this may not always work as well as in this example. One reason is that there may be semantic ambiguities which can not be resolved by syntactic considerations. Another is that the syntactic information delivered by the tagger may not be fine grained enough to resolve all syntactic ambiguities.[2] Do we need the fine grained distinctions of the large tag set to resolve ambiguities, or does the rough information from the small tag set suffice? To address these questions, we performed an evaluation using another test sample from the Frankfurter Rundschau-Corpus.

We found that - according to the Morphy lexicon - of all 9,893 word forms in the sample, 9,198 (93.0%) had an unambiguous lemma. Of the remaining 695 word forms, 667 had two possible lemmata and 28 were threefold ambiguous (Table 3 gives some examples). Using the large tag set, 616 out of the 695 ambiguous word forms were correctly lemmatized (88.6%). The corresponding figures for the small tag set were slightly better: 625 out of 695 ambiguities were resolved correctly (89.9%). When the error-rate is related to the total number of word forms in the text, the accuracy is 99.2% for the large and 99.3% for the small tag set.

The better performance when using the small tag set is somewhat surprising since there are a few cases of ambiguities in the test corpus which can only be resolved by the large tag set but not by the small tag set. For example, since the small tag set does not consider a noun's case, gender, and number, it can not decide whether *Filmen* is derived from *der Film* ("the film") or from *das Filmen* ("the filming"). On the other hand, as shown in the previous section, the tagging accuracy is much better for the small tag set, which is an advantage in lemmatization and obviously compensates for the lack of detail.

However, we believe that with future im-provements in tagging accuracy lemmatization based on the large tag set will eventually be better. Nevertheless, the current implementation of the lemmatizer gives the user the choice of selecting between either tag set.

---

[2] For example the verb *führen* can be either a subjunctive form of *fahren* ("to drive") or a regular form of *führen* ("to lead"). Since neither the large nor the small tag set consider mood, this ambiguity can not be resolved.

| Begriffen | Begriff, begreifen |
| Dank | danken, dank (prep.), Dank |
| Garten | garen, Garten |
| Trotz | Trotz, trotzen, trotz |
| Weise | Weise, weise, weisen |
| Wunder | Wunder, wundern, wund |

Table 3: Word forms with several lemmata.

## Conclusions

In this paper, a freely available integrated tool for German morphological analysis, part-of-speech tagging and context sensitive lemmatization was introduced. The morphological analyzer is based on the standard Duden grammar and provides wide coverage due to a lexicon of 324,000 word forms and the ability to process compound nouns at runtime. It gives for each word form of a text all possible lemmata and morphological descriptions. The ambiguities of the morphological descriptions are resolved by the tagger, which provides about 85% accuracy for the large and 96% accuracy for the small tag set. The lemmatizer uses the output of the tagger to disambiguate word forms with more than one possible lemma. It achieves an overall accuracy of about 99.3%.

## Acknowledgements

The work described in this paper was conducted at the University of Paderborn and supported by the Heinz Nixdorf-Institute. The Frankfurter Rundschau Corpus was generously donated by the Druck- und Verlagshaus Frankfurt am Main. We thank Gisela Zunker for her help with the acquisition and preparation of the corpus.

## Appendix: Abbreviations

| AKK | accusative | PLU | plural |
| ATT | attributive usage | POS | possessive |
| DAT | dative | PRÄ | present tense |
| FEM | feminine | PRO | pronoun |
| GEN | genitive | SIN | singular |
| IMP | imperative | SUB | noun |
| KJ1 | subjunctive 1 | SZE | punctuation mark |
| MAS | masculine | VER | verb |
| NOM | nominative | 1PE | 1st person |
| PER | personal | 3PE | 3rd person |